\pdfoutput=1
\documentclass[11pt,a4paper]{article}
\pdfoutput=1
\usepackage[hyperref]{acl2021}
\usepackage{times}
\usepackage{latexsym}

\usepackage{microtype}

\aclfinalcopy 
\def\aclpaperid{2937} 

\usepackage{xurl}
\usepackage{graphicx}
\usepackage{subfig}
\newcommand\cincludegraphics[2][]{\raisebox{-0.3\height}{\includegraphics[#1]{#2}}}
\usepackage{color, colortbl}
\definecolor{LGray}{gray}{0.9}
\definecolor{Gray}{gray}{0.8}
\definecolor{DGray}{gray}{0.7}
\usepackage{booktabs}

\title{Analyzing Online Political Advertisements}

\author{{\bf Danae S\'{a}nchez Villegas$^\alpha$} \quad {\bf Saeid Mokaram$^\beta$} \quad {\bf Nikolaos Aletras$^\alpha$}\\
    $^\alpha$ Computer Science Department, University of Sheffield, UK\\
    $^\beta$ Emotech\\
    {\small
    {\tt \{dsanchezvillegas1, n.aletras\}@sheffield.ac.uk}}\\
    {\small
    {\tt saeid.mokaram@gmail.com}}
    }

\date{}

\begin{document}
\maketitle

\begin{abstract}
Online political advertising is a central aspect of modern election campaigning for influencing public opinion. Computational analysis of political ads is of utmost importance in political science to understand the characteristics of digital campaigning. It is also important in computational linguistics to study features of political discourse and communication on a large scale.  In this work, we present the first computational study on online political ads with the aim to (1) infer the political ideology of an ad sponsor; and (2) identify whether the sponsor is an official political party or a third-party organization. We develop two new large datasets for the two tasks consisting of ads from the U.S.. Evaluation results show that our approach that combines textual and visual information from pre-trained neural models outperforms a state-of-the-art method for generic commercial ad classification. Finally, we provide an in-depth analysis of the limitations of our best-performing models and linguistic analysis to study the characteristics of political ads discourse.\footnote{Data is available here: \url{https://archive.org/details/pol_ads}}
\end{abstract}

\section{Introduction}



Online advertising is an integral part of modern digital election campaigning \cite{fulgoni2016power, fowler2020political}. The increased spending on online political ads (e.g. the 2020 U.S. election campaign spending hit an all-time record\footnote{\url{https://www.cnbc.com/2020/10/01/election-2020-campaign-spending-set-to-hit-record-11-billion.html}}) poses a significant challenge to the democratic oversight of digital campaigning,\footnote{\url{https://www.electoral-reform.org.uk/latest-news-and-research/publications/democracy-in-the-dark-digital-campaigning-in-the-2019-general-election-and-beyond/}} with serious implications about transparency and accountability, for example how voters are targeted and by whom~\citep{Kriess2020}.

Political advertising is defined as \textit{`any controlled message communicated through any channel designed to promote the political interests of individuals, parties, groups, government, or other organizations’} \cite{kaid2006television}. It is guided by ideology and morals \cite{scammell2006political,kumar2012political}, and often expresses more negativity \cite{haselmayer2019negative, iyengar1999political, lau1999effects} compared to the aesthetic nature of commercial advertising. Table \ref{t:ads_example} shows examples of online political ads across different political parties and sponsor types.

\renewcommand*{\arraystretch}{1.1}
\begin{table*}[!t]
    \footnotesize
    \centering
    \begin{tabular}{|l|l|m{10cm}|}
        \hline
        \rowcolor[HTML]{C0C0C0}
        \multicolumn{1}{|c|}{\textbf{\begin{tabular}[c]{@{}c@{}}Political\\ Ideology\end{tabular}}}
         &\textbf{ Ad Sponsor Type} & \textbf{Sample Ad}  \\
        \hline \hline
        Liberal & Political Party &\cincludegraphics[height=.6in]{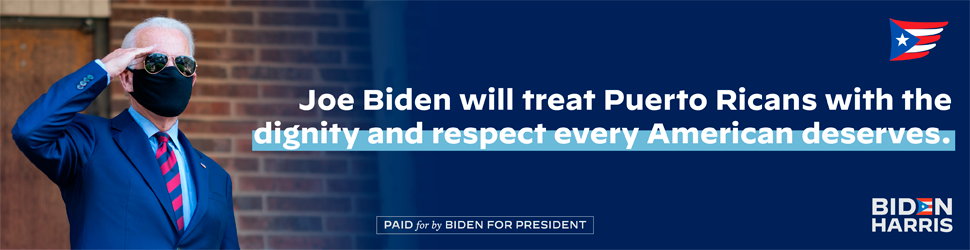}\\
         \hline \hline
        Conservative & Political Party &\cincludegraphics[height=.49in]{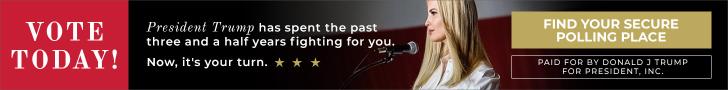}\\ 
         \hline \hline
        N/A & Third-Party &\cincludegraphics[height=.48in]{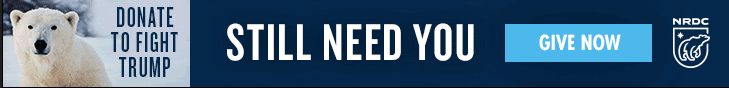} \\
        \hline
    \end{tabular}
    \caption{Examples of online political ads by sponsor political ideology and type.}
    \label{t:ads_example}
\end{table*}

While the closely related online \textit{commercial} advertising domain has recently been explored in natural language processing (NLP) for predicting the category (e.g. politics, cars, electronics) and sentiment of an ad~\cite{8099606, kalra-etal-2020-understanding}, online \textit{political} advertising has yet to be explored. Large-scale studies of online political advertising have so far focused on understanding targeting strategies rather than developing predictive models for analyzing its content \cite{edelson2019analysis,medina2020exploring}.

Automatically analyzing political ads is important in political science for researching the characteristics of online campaigns (e.g. voter targeting, sponsors, non-party campaigns, privacy, and misinformation) on a large scale \cite{scammell2006political,johansson2019analogue}. Moreover, identifying ads sponsored by third-party organizations is critical to ensuring transparency and accountability in elections \cite{liu2013adreveal, speicher2018potential,fowler2020blue,edelson2019analysis}. For example, third-party advertising had an increased presence in the U.S. House and Senate races in 2018 considerably more than in 2012 where almost half of the third-party sponsored ads were funded by dark-money sources~\citep{fowler2020blue}. Finally, computational methods for political ads analysis can help linguists to study features of political discourse and communication~\cite{kenzhekanova2015linguistic,skorupa2015linguistic}.

In this paper, we present a systematic study of online political ads (consisting of text and images) in the U.S. to uncover linguistic and visual cues across political ideologies and sponsor types using computational methods for the first time. Our contributions are as follows:
\begin{enumerate}
    \item A new classification task for predicting the political ideology (conservative or liberal) of an ad ($\S$\ref{sec:task}). We collect 5,548 distinct political ads in English from 242 different advertisers in the U.S., and label them according to the dominant political ideology of the respective sponsor's party affiliation (\emph{Liberal} or \emph{Conservative});
    \item A new classification task to automatically classify ads that were sponsored by official political parties and third-party organizations, such as businesses and non-profit organizations ($\S$\ref{sec:task}). For this task, we extract 15,116 advertisements in English from 665 distinct advertisers in the U.S., and label them as \emph{Political Party} (i.e. officially registered) and \emph{Third-Party} (i.e. other organizations) following \citet{fowler2020blue};
    \item Experiments with text-based and multimodal (text and images) models ($\S$\ref{sec:methods}) for political ideology prediction and sponsor type classification reaching up to 75.76 and 87.36 macro F1 in each task respectively ($\S$\ref{sec:results});
    \item Analysis of textual and visual features of online political ads ($\S$\ref{sec:analysis}) and error analysis to understand model limitations.
\end{enumerate}

\section{Related Work}

\subsection{Political Communication and Advertising}

Previous work on analyzing political advertising has covered television and online ads \cite{kaid2005political, reschke-anand-2012-political, west2017air, fowler2020blue}. \citet{ridout2010political} analyze a series of YouTube videos posted during the 2008 presidential campaign to understand its influence on election results as well as the actors and formats compared to traditional television ads. \citet{anstead2018political} study how online platforms such as Facebook are being used for political communication and identify challenges for understanding the role of these platforms in political elections, highlighting the lack of transparency~\citep{caplan2016controls}. \citet{fowler2020blue} explore differences between television and online ads, and demonstrate that there is a greater number of candidates advertising online than on television.

\subsection{Political Ideology Prediction}

Inferring the political ideology of various types of text including news articles, political speeches and social media has been vastly studied in NLP \cite{lin2008joint,gerrish2011predicting,sim-etal-2013-measuring,iyyer-etal-2014-political,preotiuc-pietro-etal-2017-beyond,kulkarni-etal-2018-multi, stefanov-etal-2020-predicting}. \citet{bhatia-p-2018-topic} exploit topic-specific sentiment analysis for ideology detection (i.e. conservative, liberal) in speeches from the U.S. Congress. \citet{kulkarni-etal-2018-multi} propose a multi-view model that incorporates textual and network information to predict the ideology of news articles. \citet{johnson-goldwasser-2018-classification} investigate the relationship between political ideology and language to represent morality by analyzing political slogans in tweets posted by politicians. \citet{maronikolakis-etal-2020-analyzing} present a study of political parody on Twitter focusing on the linguistic differences between tweets shared by real and parody accounts. \citet{baly-etal-2019-multi} estimate the trustworthiness and political ideology (left/right bias) of news sources as a multi-task problem. \citet{stefanov-etal-2020-predicting} develop methods to predict the overall political leaning (left, center or right) of online media and popular Twitter users.

Political ideology and communicative intents have also been studied in computer vision. Political images have been analyzed to infer the persuasive intents using various features such as facial display types, body poses, and scene context \cite{6909429, 7789592,joo2018image, bai2020m2p2, chen2020understanding}. \citet{7410780} introduce a method that infers the perceived characteristics of politicians using face images and show that those characteristics can be used in elections forecasting. \citet{xi2020understanding} analyze the political ideology of Facebook photographs shared by members of the U.S. Congress. \citet{chen2020understanding} examine the role of gender stereotypical cues from photographs posted in social media by political candidates and their relationship to voter support.

\subsection{Computational Analysis of Online Ads}
\citet{8099606} propose the task of ad understanding using vision and language. The aim is to predict the topical category, sentiment and rhetoric of an ad (i.e. what the message is about). The latter task has been approached as a visual question-answering task by ranking human generated statements that explain the intent of the ad in computer vision \cite{ye2018advise, ahuja2018understanding}. More recently in NLP, \citet{kalra-etal-2020-understanding} propose a BERT-based~\citep{devlin-etal-2019-bert} model for this task using the text and visual descriptions of the ad \cite{7780863}. \citet{thomas2018persuasive} study the persuasive cues of faces across ad categories (e.g. beauty, clothing). \citet{zhang2018equal} explore the relationship between the text of an ad and the visual content to analyze the semantics across modalities. \citet{ye2018story} integrates audio and visual modalities to predict the climax of an advertisement (i.e. stress levels) using sentiment annotations.


\renewcommand*{\arraystretch}{1.2}
\begin{table*}[!t]
    \footnotesize
    \centering
    \begin{tabular}{|l|m{10cm}|}
        \hline
         \cellcolor[HTML]{C0C0C0}Sample Ad &\cincludegraphics[height=.5in]{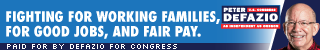}\\
         \hline \hline
        \cellcolor[HTML]{C0C0C0} Image Text & FIGHTING FOR WORKING FAMILIES, FOR GOOD JOBS, AND FAIR PAY.
        PAID FOR BY DEFAZIO FOR CONGRESS\\
         \hline \hline
        \cellcolor[HTML]{C0C0C0} Densecap & the man is wearing glasses, a man holding a red tie, the background is blue\\
        \hline
    \end{tabular}
    \caption{Example of text, and visual information extracted from a sample Ad.}
    \label{t:textex}
\end{table*}

\section{Tasks \& Data}
\label{sec:task}
We aim to analyze the political ideology of ads consisting of image and text, and the type of the ad sponsor for the first time. To this end, we present two new binary classification tasks motivated by related studies in political communication~\citep{grigsby2008,fowler2020blue}:
\begin{itemize}
    \item \textbf{Task 1: \emph{Conservative}/\emph{Liberal}} The aim is to label an ad according to the political party that sponsored the ad either as \emph{Conservative} (i.e. assuming that the dominant ideology of the Republican Party is conservatism), or \emph{Liberal} (i.e. assuming that the dominant ideology of the Democratic Party is social liberalism)~\citep{grigsby2008};
     \item \textbf{Task 2: \emph{Political Party}/\emph{Third-Party}} The goal is to classify an ad according to the type of the organization that sponsored the ad. We distinguish between ads sponsored by official political parties and non-political organizations, such as businesses and non-profit groups, following \citet{fowler2020blue}.
\end{itemize}

To the best of our knowledge, no datasets are available for modeling these two tasks. Therefore, we develop two new publicly available datasets consisting of political ads and ideology/sponsor type labels from the U.S.. We opted to use data only from the U.S. because its Federal Election Commission\footnote{\url{https://www.fec.gov/}} (FEC) provides publicly available information of political ads sponsors such as official political parties (e.g. Democratic, Republican) via their FEC ID; and third-party organizations can be identified via their Employer Identification Number\footnote{\url{https://www.irs.gov/businesses/small-businesses-self-employed/do-you-need-an-ein}} (EIN) suitable for our study.
\subsection{Collecting Online Political Ads}
We use the public Google transparency report platform\footnote{\url{https://transparencyreport.google.com/political-ads/region/US}} to collect political ads. This platform contains information on verified political advertisers (i.e. sponsors) and provides links to actual political ads from Google Ad Services.

We collect all U.S. available data from the Google platform consisting of ads published from May 31, 2018 up to October 11, 2020 (note that there is no data prior to 2018). This corresponds to a total of 168,146 image ads. Each ad is associated with a URL that links to its summary metadata consisting of a URL to the original image file and sponsor information, i.e. name and FEC ID, state elections registration or EIN ID.\footnote{All ad sponsors must apply for eligibility verification in order to publish political ads on Google platforms -  \url{https://support.google.com/displayvideo/answer/9014141}}

We scrape all available image files resulting into a total of 158,599 ads which corresponds to 94.32\% of all ads in the Google database. The rest of the ads were either not available due to violations to Google's Advertising Policy, the summary metadata was missing, or the file URL was not included in the metadata.


\subsection{Extracting Text and Visual Information}

Before, we label the ads with ideology and sponsor type, we extract two types of information from the images: (1) the text contained in each ad (Image Text; IT) using the Google Vision API;\footnote{\url{https://cloud.google.com/vision/docs/ocr}} and (2) the descriptive caption or denscap (D) of the image using the DenseCap API,\footnote{\url{https://deepai.org/machine-learning-model/densecap}} following the method proposed by \citet{kalra-etal-2020-understanding} for commercial ad classification. This way, we obtain both the actual text appearing on the ad and the textual descriptions of the ad such as entities in the images, their characteristics and relationships. Table \ref{t:textex} shows an example of an ad consisting of an image, text information and the densecap.

We use the textual and visual information to eliminate all duplicate images by comparing the URL of the image, its text and densecap. Finally, we filter out all ads that contain non-English text (i.e. IT).\footnote{\url{https://pypi.org/project/langdetect/}} This results in 15,116 unique ads from 665 unique ad sponsors.

\subsection{Labeling Ads with Political Ideology}

Our aim is to label political ads as \emph{Conservative} or \emph{Liberal} (see Task 1 description). First, we retrieve all the ad sponsors and their corresponding ads that are available in the Google Ads database. Official political committees associated with the Democratic or Republican parties are identified by their FEC ID (included in the sponsor's information in the Google database). However, the name of the political party associated with a sponsor is not available in the Google database. Thus, we query the FEC database to obtain the affiliation for all committees of the Democratic and Republican parties (e.g. Donald J. Trump for President, Inc.). Then, we compare this information with the Google database (FEC ID and exact name), to assign the corresponding affiliation to the sponsors. For example an ad sponsored by the `Donald J. Trump for President, Inc.' official committee is labeled as \emph{Republican} and subsequently as \emph{Conservative} (in a similar way we label ads for the \emph{Liberal} class).

In total, we collect 242 unique sponsors corresponding to 5,548 ads. \emph{Liberal} ads represent the 39\% of the total ads and the rest are \emph{Conservative} (61\%).

\subsection{Labeling Ads with Sponsor Type}

We first label all ads from sponsors that have an associated FEC ID in the Google database as \emph{Political Party}. These sponsors correspond to official political committees affiliated with the  Democratic or Republican parties (e.g. Biden for President).

\emph{Third-party} sponsors of political ads consist of groups not officially associated to any political party such as not-for-profit organizations (e.g. NRDC Action Fund) and businesses \cite{fowler2020blue}. This type of sponsors are identified with their EIN ID (included in the Google database). Thus, we label all ads linked to an EIN ID as \emph{Third-Party}. We collected a total of 15,116 ads where 37\% corresponds to \emph{Political Party} and 63\% corresponds to \emph{Third-Party}.


\subsection{Data Splits}
We split both datasets chronologically into train (80\%), development (10\%), and test (10\%) sets. Table \ref{tab:temp_split} shows the dataset statistics and splits for each task.

\subsection{Data Preprocessing}

\paragraph{Text} We normalize the text from the image (IT) and the densecap (D) by lower-casing, and replacing all URLs and person names with a placeholder token. To identify the person names we use the Stanford NER Tagger \cite{finkel-etal-2005-incorporating}. Also, we replace tokens that appear in less than five ads with an `unknown' token. We tokenize the text using the NLTK tokenizer \citep{bird2009natural}. Table \ref{tab:tokens} shows the average number of tokens in IT and D for each data split. 

\paragraph{Image} Each image is resized to ($300\times300$) pixels represented by red, green and blue color values. Each color channel is an integer in the range [0, 255]. The pixel values  of all images are dived by 255 to normalize them in the range [0, 1].

\begin{table}[!t]
\centering
\resizebox{0.48\textwidth}{!}{
\begin{tabular}{lrrrr}
\hline
\rowcolor[HTML]{C0C0C0}
\multicolumn{5}{c}{\textbf{T1: Liberal/Conservative}} \\
\hline
  & \textbf{Train} & \textbf{Dev} & \textbf{Test} & \textbf{Total}  \\ \toprule
C     & 2,576 (58\%)          & 369 (69\%)        & 453   (75\%) & 3,398 (61\%)  \\ \hline
L   & 1,835 (42\%)         & 165 (31\%)        & 150   (25\%)   & 2,150  (39\%)  \\ \hline
All     & 4,411 (79.5\%)        & 534 (9.6\%)       & 603 (10.9\%)        & 5,548 (100\%)   \\ \hline
Start   & 05-31-18      & 02-01-20  &  07-04-20    &    -      \\ \hline
End    &  01-30-20      & 06-30-20  &  10-10-20     &   -      \\

\hline
\rowcolor[HTML]{C0C0C0}
\multicolumn{5}{c}{\textbf{T2: Political Party/Third-Party}} \\
\hline
  & \textbf{Train} & \textbf{Dev} & \textbf{Test} & \textbf{Total} \\ \toprule
PP      & 4,663 (39\%)         & 324 (21\%)        & 561  (37\%)        & 5,548  (37\%)    \\ \hline
TP    & 7,427  (61\%)        & 1,188  (79\%)       & 953   (63\%)       & 9,568  (63\%)   \\ \hline
All     & 12,090 (80\%)        & 1,512 (10\%)       & 1,514  (10\%)       & 15,116 (100\%)     \\ \hline
Start   & 05-31-18  & 04-14-20     & 07-20-20     & -           \\ \hline
End    & 04-13-18  & 07-19-20       &  10-11-20    &    -        \\
\bottomrule

\end{tabular}
}
\caption{Data set statistics for Task 1: \emph{Conservative} (C)/ \emph{Liberal} (L), and Task 2: \emph{Political Party} (PP)/\emph{Third-Party} (TP).}
\label{tab:temp_split}
\end{table}

\begin{table}[t!]
\centering
\resizebox{0.48\textwidth}{!}{
\begin{tabular}{cccc}
\hline
\rowcolor[HTML]{C0C0C0}
\multicolumn{4}{c}{\cellcolor[HTML]{C0C0C0}\textbf{Avg. Tokens (Train/Dev/Test)}}                                                                                                          \\ \hline

\textbf{Task}              & \textbf{\begin{tabular}[c]{@{}c@{}} IT\end{tabular}} & \textbf{D} & \textbf{\begin{tabular}[c]{@{}c@{}}IT+D\end{tabular}} \\ \toprule
T1 & 17.1/16.5/17.1   & 38.3/39.9/36.9              & 55.4/56.4/54.0     \\ \hline

T2 & 16.2/17.6/19.2  & 36.7/38.9/37.2             & 52.9/56.5/56.4     \\ \bottomrule
\end{tabular}}
\caption{Average number of tokens in image text (IT), densecaps (D) and both (IT+D) for sponsor ad ideology (T1) and type (T2) prediction.}
\label{tab:tokens}
\label{tab:avgtokens}
\end{table}

\section{Predictive Models}
\label{sec:methods}
We experiment with textual, visual and multimodal models for political ad classification. 

\subsection{Linear Baselines}
As baseline models, we use logistic regression with bag of n-grams and L2 regularization using (1) the image text (LR$_{IT}$); (2) densecap (LR$_{D}$); and (3) their concatenation (LR$_{IT+D}$) for representing each ad.

\subsection{BERT} We also test three models proposed by \citet{kalra-etal-2020-understanding} for generic ad classification demonstrating state-of-the-art performance. The models are based on Bidirectional Encoder Representations from Transformers (BERT) \cite{devlin-etal-2019-bert} using a combination of the image text and the densecap. We follow a similar approach and fine-tune BERT for predicting the corresponding class in each task by adding an output dense layer for binary classification that receives the `classification' [CLS] token as input. We use three types of inputs for each ad: (1) image text (BERT$_{IT}$); (2) densecap (BERT$_{D})$; and (3) their concatenation (BERT$_{IT+D}$).

\subsection{EfficientNet} EfficientNet \cite{pmlr-v97-tan19a} is a family of Convolutional Neural Network (CNN) \cite{lecun1995convolutional} models which has achieved state-of-the-art accuracy on ImageNet \cite{5206848}. In particular, we use EfficientNet-B3 and fine-tune it on political ad classification by adding an output dense layer for each binary classification task.

\subsection{BERT+EffN} We finally test two multimodal models by combining: (1) BERT$_{IT}$ and EfficientNet (BERT$_{IT}$+EffN); and (2) BERT$_{IT+D}$ and EfficientNet (BERT$_{IT+D}$+EffN). We concatenate the text representation obtained by BERT and the visual information from EfficientNet into a $768+1536$ dimensional vector from BERT and EfficientNet respectively. This vector is then passed to an output layer for binary classification. We fine-tune the entire architecture for each task.

\begin{table}[!t]
\centering
\renewcommand{\arraystretch}{1.2}
\resizebox{0.48\textwidth}{!}{
\begin{tabular}{|l|l|l|l|}
\hline
\rowcolor[HTML]{C0C0C0}
\multicolumn{4}{|c|}{\cellcolor[HTML]{C0C0C0}\textbf{T1: Conservative/Liberal}}                                                                                                                                                             \\ \hline
\rowcolor[HTML]{C0C0C0}
\multicolumn{1}{|c|}{\cellcolor[HTML]{C0C0C0}\textbf{Model}} & \multicolumn{1}{c|}{\cellcolor[HTML]{C0C0C0}\textbf{P}} & \multicolumn{1}{c|}{\cellcolor[HTML]{C0C0C0}\textbf{R}} & \multicolumn{1}{c|}{\cellcolor[HTML]{C0C0C0}\textbf{F1}} \\
Majority                                                     & 50.00 (0.00)                                            & 37.56 (0.00)                                            & 42.90 (0.00)                                             \\ \hline
LR$_{D}$                                                     & 55.76 (0.85)                                            & 54.91 (0.89)                                            & 54.85 (1.12)                                             \\
LR$_{IT}$                                                     & 78.38 (0.70)                                            & 71.99 (0.56)                                            & 72.65 (0.73)                                             \\
LR$_{IT+D}$                                                  & 72.57 ( 1.03)                                           & 71.52 (0.62)                                            & 71.99 (0.79)                                             \\ \hline
\citet{kalra-etal-2020-understanding}                        &                                                         &                                                         &                                                          \\
\enskip BERT$_{D}$                                           & 59.40 (0.78)                                            & 57.77 (0.98)                                           & 57.64 (1.52)                                             \\
\enskip BERT$_{IT}$                                          &  72.88 (0.24)                                            & 73.46 (0.16)                                            & 73.16 (0.20)                                             \\
\enskip BERT$_{IT+D}$                                        & 78.62 (3.14)                                            & 74.08 (2.81)                                            & 75.49 (3.01)                                             \\ \hline
EfficientNet                                                 & 69.02 (3.48)                                            & 67.87 (1.23)                                           & 68.15 (1.89)                                             \\ \hline
Ours                                                         &                                                         &                                                         &                                                          \\
\enskip BERT$_{IT}$+EffN                                     & 74.99 (1.23)                                            & 72.01 (2.27)                                            & 73.02 (2.07)                                             \\
\enskip BERT$_{IT+D}$+EffN                                   & \textbf{80.24 (0.06)}                                   & \textbf{74.59 (1.70)}                                   & \textbf{75.76 (2.19)}                                    \\ \hline
\end{tabular}
}
\caption{Macro Precision (P), Macro Recall (R), and Macro F1-Score (F1) for political ideology prediction (± std. dev. for 3 runs). Best results are in bold.}
\label{tab:resultspolid}
\end{table}

\section{Experimental Setup}
\label{sec:expsetup}


We select the hyperparameters for all neural models using early stopping by monitoring the validation binary cross-entropy loss, and we estimate the class weights using the 'balanced' heuristic \cite{king2001logistic} for each task, as both datasets are imbalanced. BERT and EfficientNet models use ADAM optimizer \cite{kingma2014adam}, and experiments use 1 GPU (Nvidia V100).

\paragraph{LR} For LR we use bag of n-grams with $n = (1,3)$, $n \in$ \{(1,1),(1,2),(1,3)\} weighted by TF.IDF and L2 regularization. The average training time is 30 seconds.

\paragraph{BERT} We fine-tune  BERT for 20 epochs and choose the epoch with the lowest validation loss. We use the pre-trained base-uncased model for BERT \cite{Vaswani2017, devlin-etal-2019-bert} from HuggingFace implementation (12-layer 768-dimensional)  trained on English Wikipedia  \cite{DBLP:journals/corr/abs-1910-03771}. The maximal sequence length is 512 tokens. We fine-tune BERT for 2 epochs and learning rate $\eta=2e^{-5}$ for ideology prediction; and $\eta=1e^{-5}$ for advertiser type prediction with $\eta \in \{1e^{-5},2e^{-5},3e^{-5},4e^{-5}\}$. The average training time is 8.1 minutes.

\paragraph{EfficientNet} We use EfficientNet-B3 with Noisy-Student weights \cite{9156610}. For ideology prediction, we first freeze the layers of the EfficientNet \cite{pmlr-v97-tan19a} model and train it for 11 epochs with learning rate $\eta=1e^{-3}$ to learn the parameters of the output layer. We then unfreeze and train the whole network for another 30 epochs with $\eta=1e^{-4}$, as it has been shown that unfreezing the CNN during the latter stages of training improves the performance of the network \cite{faghri2017vse++}. For predicting the type of sponsor, we train for 45 epochs and $\eta=1e^{-2}$ keeping the EfficientNet layers frozen. Unfreezing the base model did not result into lower validation loss. We use dropout rate of 0.2 before passing the output of EfficientNet to the classification layer. The average training time is 37.8 minutes.

\paragraph{BERT+EffN} For ideology prediction, we freeze all the layers of the pre-trained models (BERT and EfficientNet) apart from the classification layer and train for 27 epochs with $\eta=1e^{-3}$. We then fine-tune BERT for 30 epochs with $\eta=1e^{-5}$. For sponsor type prediction, we freeze all EfficientNet layers and fine-tune BERT for 30 epochs with $\eta=2e^{-6}$. We train in stages to ensure that the parameters of each part of the model (textual and visual) are properly updated \cite{kiela2019supervised}. The average training time is 56.65 minutes.

\begin{table}[!t]
\centering
\renewcommand{\arraystretch}{1.16}
\resizebox{0.48\textwidth}{!}{
\begin{tabular}{|l|l|l|l|}
\hline
\rowcolor[HTML]{C0C0C0}
\multicolumn{4}{|c|}{\cellcolor[HTML]{C0C0C0}\textbf{T2: Political Party/Third-Party}}                                                                                                                                                      \\ \hline
\rowcolor[HTML]{C0C0C0}
\multicolumn{1}{|c|}{\cellcolor[HTML]{C0C0C0}\textbf{Model}} & \multicolumn{1}{c|}{\cellcolor[HTML]{C0C0C0}\textbf{P}} & \multicolumn{1}{c|}{\cellcolor[HTML]{C0C0C0}\textbf{R}} & \multicolumn{1}{c|}{\cellcolor[HTML]{C0C0C0}\textbf{F1}} \\
Majority                                                     & 50.00 (0.00)                                            & 31.47 (0.00)                                            & 38.62 (0.00)                                             \\ \hline
LR$_{D}$                                                     & 53.60 (0.72)                                            & 53.40 (0.65)                                            & 53.11 (0.58)                                             \\
LR$_{IT}$                                                     & 84.02 (0.14)                                            & 85.04 (0.31)                                            & 84.47 (0.18)                                             \\
LR$_{IT+D}$                                                  & 86.46 (0.13)                                            & 86.63 (0.09)                                            & 86.54 (0.05)                                             \\ \hline
\citet{kalra-etal-2020-understanding}                        &                                                         &                                                         &                                                          \\
\enskip BERT$_{D}$                                           & 56.50 (0.89)                                            & 56.31 (0.78)                                            & 53.45 (1.26)                                             \\
\enskip BERT$_{IT}$                                          & 85.57 (0.86)                                            & 86.42 (2.01)                                            & 85.86 (1.23)                                             \\
\enskip BERT$_{IT+D}$                                        & 87.00 (0.89)                                   & 86.81 (0.83)                                            & 86.90 (0.86)                                             \\ \hline
EfficientNet                                                 & 53.27 (2.86)                                            & 53.93 (2.40)                                            & 51.53 (5.46)                                             \\ \hline
Ours                                                         &                                                         &                                                         &                                                          \\
\enskip BERT$_{IT}$+EffN                                   & \textbf{87.02 (2.74)}                                            & 85.81 (0.20)                                            & 86.29 (1.11)                                             \\
\enskip BERT$_{IT+D}$+EffN                &                   86.78 (0.03)                                            & \textbf{88.18 (1.10)}                                   & \textbf{87.36 (0.39)}                                    \\ \hline
\end{tabular}
}
\caption{Macro Precision (P), Macro Recall (R), and Macro F1-Score (F1) for sponsor type prediction (± std. dev. for 3 runs). Best results are in bold.}
\label{tab:resultssponsor}
\end{table}
\section{Results}
\label{sec:results}
This section presents the experimental results for the two predictive tasks, political ideology and sponsor type prediction ($\S$\ref{sec:task}) using the methods described in $\S$\ref{sec:methods}. We evaluate our models using macro precision, recall and F1 score since the data in both tasks is imbalanced. Note that for all models we report the average and standard deviation over three runs using different random seeds. We also report the majority class baseline for each task.

\begin{figure*}[t!]
\small
\centering
\begin{tabular}{cccc}
\subfloat[True: \emph{Lib} - Pred: \emph{Cons}]{\includegraphics[width = 1.4in]{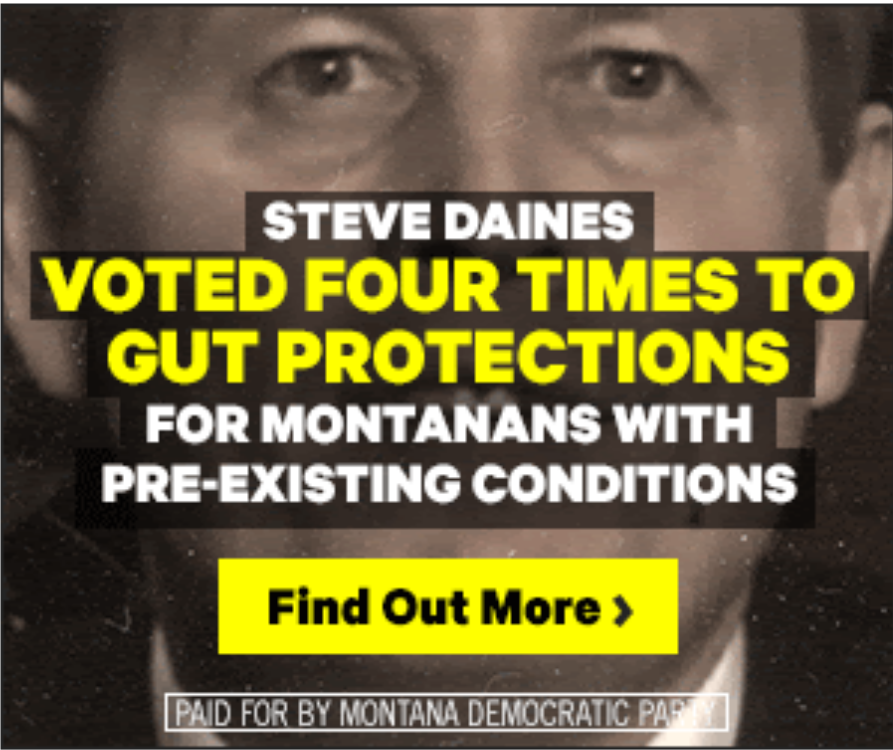}}&
\subfloat[True: \emph{Cons} - Pred: \emph{Lib}]{\includegraphics[width = 1.4in]{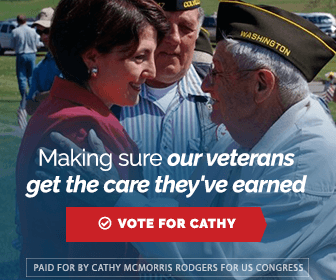}}&
\subfloat[True: \emph{PP} - Pred: \emph{TP}]{\includegraphics[width = 1.4in]{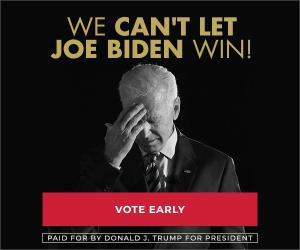}}&
\subfloat[True: \emph{TP} - Pred: \emph{PP}]{\includegraphics[width = 1.4in]{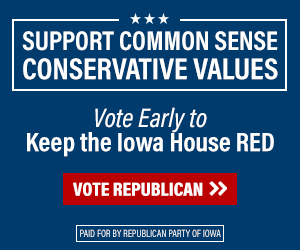}}
 \\

\end{tabular}
\caption{Examples of ads with their true and predicted labels Lib (Liberal), Cons (Conservative), PP (Political Party), TP (Third-Party).}
\label{fig:ext2}
\end{figure*}

\subsection{Predictive Performance}
\label{sec:discussion}
\paragraph{Task 1: Conservative/Liberal}
Table \ref{tab:resultspolid} shows the results for the political ideology prediction. We first observe that BERT$_{IT}$ (73.16\%) which uses as input information the image text outperforms BERT$_{D}$ (57.64\%) and EfficientNet (68.15\%) in macro F1. This suggests that the text shown on a political ad is the dominant medium for conveying its main message, corroborating findings in related research on commercial ads \cite{dey2019beyond, kalra-etal-2020-understanding}.

Moreover, combining image text and densecap (BERT$_{IT+D}$), leads to higher performance, than using only image text (BERT$_{IT}$), i.e. 75.49\% and 73.16\% F1 respectively. This indicates that the combination of textual with visual information (in the form of image descriptions) improves the model performance.

Finally, using all visual information sources, i.e. densecaps and image representation from EfficientNet (BERT$_{IT+D}$+EffN), further improves performance achieving the highest macro F1 (75.76\%) across models, followed by BERT$_{IT+D}$ (75.49\%). 

\paragraph{Task 2: Political-Party/Third-Party}
Table \ref{tab:resultssponsor} shows the results for the sponsor type prediction. The best overall performance is obtained by BERT$_{IT+D}$+EffN (87.36\%) which combines both image and textual information. BERT$_{IT+D}$ (86.90\%) and LR$_{IT+D}$ (86.54\%) follow very closely. By inspecting our data, we identified the presence of noise in image text, particularly sentences are interrupted by logos and other aesthetic elements. This negatively affects the performance of BERT because such models are usually pre-trained on `cleaner' generic corpora~\cite {kumar-etal-2020-noisy}. On the other hand, LR models trained from scratch can adapt to the noisy text (see $\S$ \ref{sec:error} for error analysis).



Overall, our results in both tasks suggest that text is a stronger modality for inferring the political ideology and sponsor type of political ads compared to visual information extracted from the images. However, integrating visual information in the form of text descriptions (densecaps) or representations obtained by pre-trained image classification models, enhances model performance.

\subsection{Error Analysis}
\label{sec:error}
We further perform an error analysis to examine the behavior of our best performing models (BERT$_{IT+D}$+EffN and BERT$_{IT+D}$) and identify potential limitations.


The ad shown in Fig.~\ref{fig:ext2} (a) was misclassified as \emph{Conservative} by BERT$_{IT+D}$ and BERT$_{IT+D}$+EffN. This particular ad requires common knowledge of social issues (e.g. inadequate health support) that are often discussed in political campaigns to inform voters about a party's views on the issue \cite{scammell2006political}. This makes the classification task difficult for the models since it requires contextual knowledge. Incorporating external relevant knowledge to the models (e.g. political speeches, interviews or public meetings) might improve performance \cite{8508244}.

The ad depicted in Fig.~\ref{fig:ext2} (b) was misclassified by BERT$_{IT+D}$ and BERT$_{IT+D}$+EffN as \emph{Conservative}. After analyzing the densecap descriptions, we found that this information tends to be noisy. For this particular example, it contains descriptions such as `a man is holding a horse', `the sign is blue', `a blue and white stripe shirt', and `a man wearing a hat'. In fact, BERT$_{IT}$, which only takes the image text into account, classified this ad correctly as \emph{Conservative}. Improving the quality of the image descriptions (e.g. pre-training on advertising or political images, capturing specific attributes such as `military hat') might be beneficial for these models.

Fig.~\ref{fig:ext2} (c) shows an example of a \emph{Political Party} ad misclassified by BERT$_{IT+D}$+EffN as \emph{Third-Party}. The ad contains the following text:
\begin{quote}
    \small
    \it
    WE CAN'T LET $<$person$>$ WIN!

    VOTE EARLY
\end{quote}

The message has a confrontational and divisive tone that is common in \emph{Third Party} ads \cite{edelson2019analysis}, but is typically used as a political tactic for negative campaigning \cite{skaperdas1995modeling, gandhi2016negative, haselmayer2019negative}. 

Finally, Fig. \ref{fig:ext2} (d) shows an example of a \emph{Third-Party} ad misclassified as \emph{Political Party} by BERT$_{IT+D}$+EffN. The text content promotes voter participation (e.g. \textit{Vote}), a characteristic of \emph{Political Party} advertising (see Table \ref{tab:lat2}). However, one of the aims of the \emph{Third-Party} advertising is precisely to encourage voting and activism \cite{dommett2018digital}.

There is a considerable difference between the models using visual information only (LR$_D$, BERT$_D$, EfficientNet), and those that also use the ad text as input (IT, IT+D). Our intuition is that models get confused by the appearance of shapes, colors and other aesthetic features that are domain specific and appear frequently in political advertisements \cite{sartwell2011political}. For instance, several ads that belong to the \emph{Third-Party} category, include buttons linking to websites (see Fig, \ref{fig:ext2} (c), (d)). However, \emph{Political Party} ads, also make use of these type of buttons to link users to donation or informative websites \cite{edelson2019analysis}.







\begin{table}[t!]
\small\centering
\begin{tabular}{|l|l|l|l|}
\hline
\rowcolor[HTML]{9B9B9B}
\multicolumn{2}{|c|}{\cellcolor[HTML]{9B9B9B}\textbf{Liberal}}                                                                                                         & \multicolumn{2}{c|}{\cellcolor[HTML]{9B9B9B}\textbf{Conservative}}                                                      \\ \hline
\rowcolor[HTML]{9B9B9B}
\multicolumn{1}{|c|}{\cellcolor[HTML]{9B9B9B}\textbf{\begin{tabular}[c]{@{}c@{}}Feature\\ \end{tabular}}} & \multicolumn{1}{c|}{\cellcolor[HTML]{9B9B9B}\textbf{r}} & \multicolumn{1}{c|}{\cellcolor[HTML]{9B9B9B}\textbf{Feature}} & \multicolumn{1}{c|}{\cellcolor[HTML]{9B9B9B}\textbf{r}} \\ \hline
necessary                                                                                                    & 0.197                                                   & senate                                                        & 0.271                                                   \\ \hline
end                                                                                                         & 0.196                                                   & republican                                                    & 0.196                                                   \\ \hline
prohibited                                                                                                   & 0.190                                                   & !                                                             & 0.176                                                   \\ \hline
approx                                                                                                       & 0.186                                                   & conservative                                                  & 0.127                                                   \\ \hline
contrib                                                                                                      & 0.181                                                   & national                                                      & 0.116                                                   \\ \hline
void                                                                                                         & 0.177                                                   & committee                                                     & 0.112                                                   \\ \hline
values                                                                                                        & 0.173                                                   & petition                                                      & 0.109                                                   \\ \hline
prz                                                                                                          & 0.161                                                   & border                                                        & 0.102                                                   \\ \hline
subj                                                                                                         & 0.156                                                   & taxes                                                         & 0.099                                                   \\ \hline
make                                                                                                         & 0.156                                                   & radical                                                       & 0.098                                                   \\ \hline
win                                                                                                          & 0.144                                                   & sign                                                          & 0.096                                                   \\ \hline
place                                                                                                        & 0.140                                                   & stop                                                          & 0.094                                                   \\ \hline
beer                                                                                                         & 0.139                                                   & states                                                        & 0.093                                                   \\ \hline
\end{tabular}
\caption{Feature correlations with \emph{Conservative}/\emph{Liberal} Ads, sorted by Pearson correlation (r). All correlations are significant at $p < .01$, two-tailed t-test.}
\label{tab:lat1}
\end{table}
\section{Linguistic Analysis}
\label{sec:analysis}
We perform an analysis based on our new data set to study the linguistic characteristics of political ads. We first analyze the specific features of each class for both tasks. For this purpose, we use a method introduced by~\citet{schwartz2013personality} to analyze uni-gram features from image text (see $\S$\ref{sec:methods}) using univariate Pearson correlation. Features are normalized to sum up to unit for each ad. For each feature, we compute correlations independently between its distribution across ads and its label (\emph{Conservative}/\emph{Liberal}), or \emph{Political Party}/\emph{Third Party}).

\subsection{Conservative vs. Liberal}
Table \ref{tab:lat1} presents the top unigrams correlated with \emph{Liberal} and \emph{Conservative} ads. We first notice that the top words in the \emph{Conservative} category are closely related to its ideology such as `conservative' and `republican'. Other prominent terms in these categories are words related to current political issues, such as immigration (e.g. `border') and taxation (e.g. `taxes'). In fact, these are examples of emotionally evocative terms (e.g. anger about taxes) that are frequently used in political campaigns to influence voters \cite{brader2005striking}.

Top terms of \emph{Liberal} ads include `necessary', `end',`values', and `win'. For example, the following ads belong to the \emph{Liberal} class:
\begin{quote}
\small
\it
    I'm supporting $<$person$>$ because he has the same \textbf{values} that I do and he's an honest person.
\end{quote}

\begin{quote}
    \small
    \it
     $<$person$>$ FOR CONGRESS

     To \textbf{End} Gun Violence
\end{quote}
These are examples of ads containing a combination of moral and controversial topics (e.g. gun regulation) which are typical characteristics of political advertising \cite{kumar2012political}.

\begin{table}[t!]
\small
\centering
\begin{tabular}{|r|r|l|r|}
\hline
\rowcolor[HTML]{9B9B9B}
\multicolumn{2}{|c|}{\cellcolor[HTML]{9B9B9B}\textbf{\begin{tabular}[c]{@{}c@{}}Political Party\end{tabular}}}                                                       & \multicolumn{2}{c|}{\cellcolor[HTML]{9B9B9B}\textbf{\begin{tabular}[c]{@{}c@{}}Third-Party\end{tabular}}}             \\ \hline
\rowcolor[HTML]{9B9B9B}
\multicolumn{1}{|c|}{\cellcolor[HTML]{9B9B9B}\textbf{\begin{tabular}[c]{@{}c@{}}Feature\end{tabular}}} & \multicolumn{1}{c|}{\cellcolor[HTML]{9B9B9B}\textbf{r}} & \multicolumn{1}{c|}{\cellcolor[HTML]{9B9B9B}\textbf{Feature}} & \multicolumn{1}{c|}{\cellcolor[HTML]{9B9B9B}\textbf{r}} \\ \hline
congress                                                                                                     & 0.365                                                   & state                                                         & 0.193                                                   \\ \hline
vote                                                                                                         & 0.308                                                   & learn                                                         & 0.181                                                   \\ \hline
senate                                                                                                       & 0.292                                                   & champion                                                      & 0.175                                                   \\ \hline
!                                                                                                            & 0.269                                                   & senator                                                       & 0.166                                                   \\ \hline
president                                                                                                    & 0.248                                                   & thank                                                         & 0.153                                                   \\ \hline
committee                                                                                                    & 0.236                                                   & action                                                        & 0.147                                                   \\ \hline
candidate                                                                                                    & 0.223                                                   & congressman                                                   & 0.130                                                   \\ \hline
republican                                                                                                   & 0.208                                                   & urge                                                          & 0.129                                                   \\ \hline
authorized                                                                                                   & 0.208                                                   & protect                                                       & 0.128                                                   \\ \hline
donate                                                                                                       & 0.202                                                   & access                                                        & 0.119                                                   \\ \hline
join                                                                                                         & 0.199                                                   & award                                                         & 0.117                                                   \\ \hline
\textless{}url\textgreater{}                                                                                 & 0.187                                                   & american                                                      & 0.116                                                   \\ \hline
\$                                                                                                           & 0.180                                                   & ?                                                             & 0.113                                                   \\ \hline
\end{tabular}
\caption{Feature correlations with Political Party/\emph{Third-Party} Ads, sorted by Pearson correlation (r). All correlations are significant at $p < .01$, two-tailed t-test.}
\label{tab:lat2}
\end{table}

\subsection{Political Party vs. Third-Party}
Table \ref{tab:lat2} shows the top unigram features correlated with the sponsor type of an ad (\emph{Political Party}/\emph{Third-Party}). We observe that some top terms in the \emph{Political Party} class also belong to the top terms of the political ideology task (see Table \ref{tab:lat1}) such as `committee', `republican' and `senate'. Messages calling for vote and donation support (`vote', `donate', `\$') are also prevalent in \emph{Political Party} ads \cite{fulgoni2016power}, as in the next example (See Fig. \ref{fig:ext2} (b)):
\begin{quote}
    \small
    \it
    Making sure our veterans

    get the care they've earned

    \textbf{VOTE} FOR $<$person$>$
\end{quote}

On the other hand, top features from the \emph{Third-Party} category (e.g. `action', `protect') share common characteristics with the rhetoric used by media outlets focused on promoting specific political messaging \cite{edelson2019analysis,  dommett2018digital}. Many of these ads direct people to websites to read about a particular topic. For example:

\begin{quote}
\small
\it
    Is $<$person$>$ HIDING ANTI-GUN VIEWS\textbf{?} \textbf{Learn} More
\end{quote}

This ad belongs to the \emph{Third-Party} class and points the viewer to an external website for reading further details.

\section{Conclusion}
We have presented the first study in NLP for analyzing the language of political ads motivated by prior studies in political communication. We have introduced two new publicly available datasets containing political ads from the U.S. in English labeled by (1) the ideology of the sponsor (\emph{Conservative}/\emph{Liberal}); and (2) the sponsor type (\emph{Political Party}/\emph{Third Party}). We have defined both tasks as advertisement-level binary classification and evaluated a variety of approaches, including textual, visual and multimodal models reaching up to 75.76 and 87.36 macro F1 in each task respectively. 

In the future, we aim to incorporate other modalities such as speech, and video, and explore other methods of acquiring and integrating multimodal information. In addition, we aim to extend our work for analyzing political advertising discourse across different regions, languages and platforms.


\section*{Acknowledgments}

We would like to thank Kate Dommett, Alexandra Boutopoulou, Mali Jin, Katerina Margatina, George Chrysostomou, Peter Vickers, Emily Lau, and all reviewers for their valuable feedback. DSV is supported by the Centre for Doctoral Training in Speech and Language Technologies (SLT) and their Applications funded by the UK Research and Innovation grant EP/S023062/1. NA is supported by a Leverhulme Trust Research Project Grant.

\section*{Ethics Statement}
Our work complies with the Terms of Service of the Google Political Ads Dataset.\footnote{\url{https://console.cloud.google.com/marketplace/product/transparency-report/google-political-ads?pli=1}} We provide, for reproducibility purposes, the list of ad IDs and corresponding labels used for each task, as well as the data splits (train, development, test). All data used in this paper is in English. The ads information can be retrieved from Google according to their policy.
\bibliographystyle{acl_natbib}
\bibliography{anthology,polAds}

\end{document}